\documentclass[10pt,twocolumn,letterpaper]{article}

\usepackage{wacv}
\usepackage{times}
\usepackage{epsfig}
\usepackage{graphicx}
\usepackage{amsmath}
\usepackage{amssymb}

\usepackage{listings}
\usepackage{xcolor}

\definecolor{codegreen}{rgb}{0,0.6,0}
\definecolor{codegray}{rgb}{0.5,0.5,0.5}
\definecolor{codepurple}{rgb}{0.58,0,0.82}
\definecolor{backcolour}{rgb}{0.95,0.95,0.92}

\lstdefinestyle{mystyle}{
  backgroundcolor=\color{backcolour},   commentstyle=\color{codegreen},
  keywordstyle=\color{magenta},
  numberstyle=\tiny\color{codegray},
  stringstyle=\color{codepurple},
  basicstyle=\ttfamily\footnotesize,
  breakatwhitespace=false,         
  breaklines=true,                 
  captionpos=b,                    
  keepspaces=true,                 
  numbers=left,                    
  numbersep=5pt,                  
  showspaces=false,                
  showstringspaces=false,
  showtabs=false,                  
  tabsize=2
}

    \usepackage{relsize}
    \usepackage{graphicx}
    \usepackage{subcaption}
    \usepackage{amsmath}
    \usepackage{mathtools}
    \usepackage{algorithm}
    \usepackage[noend]{algpseudocode}
    \usepackage{xcolor,colortbl}
    \usepackage{amssymb}
    \usepackage{moresize}
    \usepackage{multirow}
    \usepackage{mdframed}
    \usepackage{stmaryrd}
    \usepackage{setspace}
    \usepackage{enumitem}
    \usepackage{comment}
    
    \newcounter{example}[section]
    \newenvironment{example}[1][]{\refstepcounter{example}\par\medskip
    \noindent \textbf{Example~\theexample} \rmfamily}{\medskip}
    
    \newcounter{definition}[section]
    \newenvironment{definition}[1][]{\refstepcounter{definition}\par\medskip
    \noindent \textbf{Definition~\thedefinition~(#1)} \rmfamily}{\medskip}
    
    \newcounter{proposition}[section]

    \newcounter{theorem}[section]
    \newenvironment{theorem}[1][]{\refstepcounter{theorem}\par\medskip
    \noindent \textbf{Theorem~\thetheorem} \rmfamily}{\medskip}

    \usepackage{prettyref}
    \newrefformat{def}{Definition~\ref{#1}}
    \newrefformat{prop}{Proposition~\ref{#1}}
    \newrefformat{th}{Theorem~\ref{#1}}
    \newrefformat{algo}{Algorithm~\ref{#1}}
    \newrefformat{fig}{Figure~\ref{#1}}
    \newrefformat{tab}{Table~\ref{#1}}
    \newrefformat{ex}{Example~\ref{#1}}
    \newcommand{\pref}{\prettyref}
    
    \newcommand{\refli}[1]{line~\ref{#1}}

    \definecolor{gray50}{gray}{0.45}

    \newcommand{\ignore}[1]{}
    \newcommand{\resp}{resp.\ }

    \newcommand{\N}{\ensuremath{\mathbb{N}}}
    \newcommand{\segm}[2]{\ensuremath{\llbracket #1 ; #2 \rrbracket}}
    \newcommand{\card}[1]{\ensuremath{|#1|}}
    \newcommand{\powerset}{\ensuremath{\wp}}
    
    \newcommand{\mvl}{\ensuremath{\mathcal{M}\mathrm{V}\mathrm{L}}}
    \newcommand{\mvlp}{\ensuremath{\mathcal{M}\mathrm{V}\mathrm{L}\mathrm{P}}}
    \newcommand{\V}{\ensuremath{\mathcal{V}}}
    \newcommand{\var}{\ensuremath{\mathrm{v}}}
    
    \newcommand{\val}{\ensuremath{{val}}}
    \newcommand{\vvi}[1]{\ensuremath{\var_{#1}^{\val_{#1}}}}
    \newcommand{\vv}{\ensuremath{\var^{\val}}}
    
    \newcommand{\hvar}[1]{\ensuremath{\mathrm{var}({h({#1})})}}
    \newcommand{\bvar}[1]{\ensuremath{\mathrm{var}({b({#1})})}}
    \newcommand{\setvar}[1]{\ensuremath{\mathrm{var}({#1})}}
    \newcommand{\dom}{\ensuremath{\mathsf{dom}}}
    \newcommand{\A}{\ensuremath{\mathcal{A}}}
    \newcommand{\Sall}{\ensuremath{\mathcal{S}}}
    
    \newcommand{\dmvlp}{\ensuremath{\mathcal{D}\mathcal{M}\mathrm{V}\mathrm{L}\mathrm{P}}}
    \newcommand{\F}{\ensuremath{\mathcal{F}}}
    \newcommand{\T}{\ensuremath{\mathcal{T}}}
    \newcommand{\Val}{\ensuremath{\mathcal{V}al}}
    \newcommand{\SallF}{\ensuremath{\Sall^{\mathcal{F}}}}
    \newcommand{\SallT}{\ensuremath{\Sall^{\mathcal{T}}}}

\def\wacvPaperID{09} 

\wacvfinalcopy 

\ifwacvfinal
\fi

\ifwacvfinal
\usepackage[breaklinks=true,bookmarks=false]{hyperref}
\else
\usepackage[pagebackref=true,breaklinks=true,colorlinks,bookmarks=false]{hyperref}
\fi

\ifwacvfinal
\else
\fi

\begin{document}

\title{Symbolic AI for XAI: Evaluating LFIT Inductive Programming \\for Fair and Explainable Automatic Recruitment}

\author{Alfonso Ortega, Julian Fierrez, Aythami Morales, Zilong Wang
\\
School of Engineering, Universidad Autonoma de Madrid\\
{\tt\small \{alfonso.ortega,julian.fierrez,aythami.morales\}@uam.es, zilong.wang@estudiante.uam.es} 
\and
Tony Ribeiro \\
Laboratoire des Sciences du Num\'erique de Nantes
National Institute of Informatics
Japan \\
{\tt\small tony\_ribeiro@ls2n.fr}\\

}

\maketitle
\thispagestyle{empty}

\begin{abstract}
    Machine learning methods are growing in relevance for biometrics and personal information processing in domains such as forensics, e-health, recruitment, and e-learning. In these domains, white-box (human-readable) explanations of systems built on machine learning methods can become crucial. Inductive Logic Programming (ILP) is a subfield of symbolic AI aimed to automatically learn declarative theories about the process of data. Learning from Interpretation Transition (LFIT) is an ILP technique that can learn a propositional logic theory equivalent to a given black-box system (under certain conditions). The present work takes a first step to a general methodology to incorporate accurate declarative explanations to classic machine learning by checking the viability of LFIT in a specific AI application scenario: fair recruitment based on an automatic tool generated with machine learning methods for ranking Curricula Vitae that incorporates soft biometric information (gender and ethnicity). We show the expressiveness of LFIT for this specific problem and propose a scheme that can be applicable to other domains.

\end{abstract}

\section{Introduction}


\begin{figure*}[t]
\begin{center}
        \includegraphics[width=0.95\linewidth]{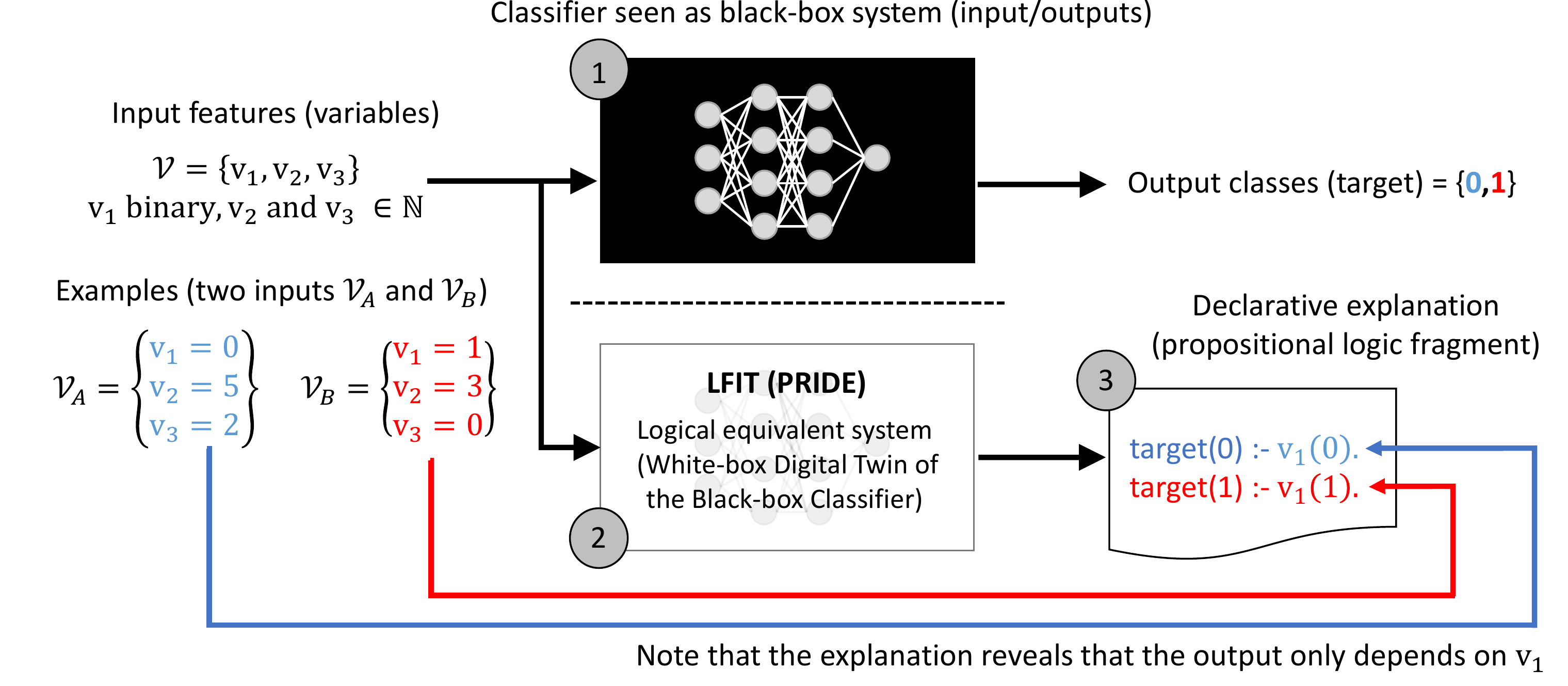}
\end{center}
   \caption{Architecture of the proposed approach for generating an explanation of a given black-box Classifier (1) using \textbf{PRIDE} (2) with a toy example (3). Note that the resulting explanations generated by \textbf{PRIDE} are in propositional logic.}
\label{fig:layout}
\label{fig:onecol}
\end{figure*}

Statistical and optimization-based machine learning algorithms have achieved great success in various applications such as speech recognition \cite{SeniorAl2012}, image classification \cite{imagenet_cvpr09}, machine translation \cite{WuAl2016}, and so on. Among these approaches, deep neural networks have shown the most remarkable success, especially in speech and image recognition. Although deep learning methods usually have good generalization ability on similarly distributed new data, they have some weaknesses including the lack of explanations and the poor understandability by humans of the whole learning process. A deep review about this question can be found in \cite{HerreraAl20}.

Another characteristic of these machine learning algorithms is that they rely on data, and therefore reflect those data. This could be an advantage in general, but in some particular domains it could be an important drawback. Consider, for example, automatic recruitment systems, or algorithms to authorize financial products. In these domains, ethic behavior is mandatory and biased data are unacceptable. Appropriate measures have to be taken for guaranteeing ethical AI behavior sometimes contradictory to the possibly biased training data. These questions are receiving increasing interest \cite{AcienAl2018, DrozdowskiAl20, NagpalAlt19, 2020_AAAI_Discrimination_Serna,
2020_ICPR_InsideBias_Serna,
2021_TPAMI_SensitiveNets_Morales}. 

On the other hand, logic programming is a declarative programming paradigm with a high level of abstraction. It is based on a formal model (first order logic) to represent human knowledge. Inductive Logic Programming (ILP) has been developed for inductively learning logic programs from examples \cite{Muggleton91}. Roughly speaking, given a collection of positive and negative examples and background knowledge, ILP systems learn declarative (symbolic) programs \cite{MuggletonAl14, CropperAl19}, which could even be noise tolerant \cite{DaiAl15, MuggletonAl18b}, that entail all of the positive examples but none of the negative examples. 

One of the ILP most promising approaches for us is Learning From Interpretation Transition (\textbf{LFIT}) \cite{Ribeiro15}. \textbf{LFIT} learns a logic representation (digital twin) of dynamical complex systems by observing their behavior as a black box under some circumstances. The most general of \textbf{LFIT} algorithms is \textbf{GULA} (General Usage LFIT Algorithm). \textbf{PRIDE} is an approximation to \textbf{GULA} with polynomial performance. \textbf{GULA} and \textbf{PRIDE} generate a propositional logic program equivalent to the system under consideration. These approaches will be introduced in depth in the following sections.

Figure~\ref{fig:layout} shows the architecture of our proposed approach for generating white-box explanations using \textbf{PRIDE} of a given black-box classifier.

The main contributions of this work are:
\begin{itemize}
\item We have proposed a method to provide declarative explanations using \textbf{PRIDE} about the classification process made by an algorithm automatically learnt by a neural architecture in a typical machine learning scenario. Our approach guarantees the logical equivalence between the explanations and the algorithm with respect to the data used to feed \textbf{PRIDE}. It does not depend on any particular characteristic of the domain, so it could be applied to any problem.
\item We have checked the expressive power of these explanations by experimenting in a multimodal machine learning testbed around automatic recruitment including different biases (by gender and ethnicity).
\end{itemize}

The rest of the paper is structured as follows: Sec.~\ref{stateofart} summarizes the related relevant literature. Sec.~\ref{sec:methods} describes our methodology including \textbf{LFIT}, \textbf{GULA}, and \textbf{PRIDE}. Sec.~\ref{experiments} presents the experimental framework including the datasets and experiments conducted. Sec.~\ref{results} presents our results. Finally Secs.~\ref{discussion}, \ref{conclusions} and \ref{further} respectively discuss our work and describe our conclusions and further research lines.

\section{Related Works: Inductive Programming for XAI \label{stateofart}}


Some meta-heuristics approaches (as genetic algorithms) have been used to automatically generate programs. Genetic programming (GP) was introduced by Koza \cite{Koza92} for automatically generating LISP expressions for given tasks expressed as pairs input/output. This is, in fact, a typical machine learning scenario. GP was extended by the use of formal grammars to allow to generate programs in any arbitrary language keeping not only syntactic correctness \cite{NeillAl03} but also semantic properties \cite{OrtegaAl07}. Algorithms expressed in any language are declarative versions of the concepts learnt what makes evolutionary automatic programming algorithms machine learners with good explainability.

Declarative programming paradigms (functional, logical) are as old as computer science and are implemented in multiple ways, e.g.: LISP \cite{Steele90}, Prolog \cite{Bratko12}, Datalog \cite{HuangGL11}, Haskell \cite{Thompson11}, and Answer Set Programs (ASP) \cite{GebserAl12}.


Of particular interest for us within declarative paradigms is logic programming, and in particular first order logic programming, which is based on the Robinson's resolution inference rule that automates the reasoning process of deducing new clauses from a first order theory \cite{Lloyd87}. Introducing examples and counter examples and combining this scheme with the ability of extending the initial theory with new clauses it is possible to automatically induce a new theory that (logically) entails all of the positive examples but none of the negative examples. The underlying theory from which the new one emerges is considered \textit{background knowledge}. This is the hypothesis of Inductive Logic Programming (ILP, \cite{MuggletonAl90,CropperAl19}) that has received a great research effort in the last two decades. 
Recently, these approaches have been extended to make them noise tolerant (in order to overcome one of the main drawbacks of ILP vs statistical/numerical approaches when facing bad-labeled or noisy examples \cite{MuggletonAl18b}). 

Other declarative paradigms are also compatible with ILP, e.g., MagicHaskeller (that implements \cite{Katayama05}) with the functional programming language Haskell, and Inductive Learning of Answer Set Programs (ILASP) \cite{Law18}. 


It has been previously mentioned that ILP implies some kind of \textit{search} in spaces that can become huge. This search can be eased by hybridising with other techniques, e.g., \cite{Alireza13} introduces GA-Progol that applies evolutive techniques. 

Within ILP methods we have identified \textbf{LFIT} as specially relevant for explainable AI (XAI). In the next section we will describe the fundamentals of \textbf{LFIT} and its \textbf{PRIDE} implementation, which will be tested experimentally for XAI in the experiments that will follow.

\subsection{Learning From Interpretation Transition (LFIT)}

    
    Learning From Interpretation Transition (\textbf{LFIT}) \cite{LFIT2013} has been proposed to automatically construct a model of the dynamics of a system from the observation of its state transitions.
    Given some raw data, like time-series data of gene expression, a discretization of those data in the form of state transitions is assumed.
    From those state transitions, according to the semantics of the system dynamics, several inference algorithms modeling the system as a logic program have been proposed.
    The semantics of a system's dynamics can indeed differ with regard to the synchronism of its variables, the determinism of its evolution and the influence of its history.
    
    
    The \textbf{LFIT} framework proposes several modeling and learning algorithms to tackle those different semantics.
    To date, the following systems have been tackled: memory-less deterministic systems
    \cite{LFIT2013}, systems with memory \cite{TRFrontier15}, probabilistic systems \cite{DMTRICLP15} and their multi-valued extensions \cite{TRICMLA15,DMTRICAPS16}.
    The work \cite{TRILP2017} proposes a method that allows to deal with continuous time series data, the abstraction itself being learned by the algorithm.
    
    In \cite{TRILP18,TRMLJ2020}, \textbf{LFIT} was extended to learn systems dynamics independently of its update semantics.
    That extension relies on a modeling of discrete memory-less multi-valued systems as logic programs in which each rule represents that a variable possibly takes some value at the next state, extending the formalism introduced in \cite{LFIT2013,TRILP14}.
    The representation in \cite{TRILP18,TRMLJ2020} is based on annotated logics \cite{Blair1989135,blair1988paraconsistent}.
    Here, each variable corresponds to a domain of discrete values.
    In a rule, a literal is an atom annotated with one of these values.
    It allows to represent annotated atoms simply as classical atoms and thus to remain at a propositional level.
    This modeling allows to characterize optimal programs independently of the update semantics.
    It allows to model the dynamics of a wide range of discrete systems including our domain of interest in this paper. {\bf LFIT} can be used to learn an equivalent propositional logic program that provides explanation for each given observation.
    

\section{Methods}
\label{sec:methods}
\subsection{General Methodology}
\begin{figure}[t]
\begin{center}
        \includegraphics[width=1.0\linewidth]{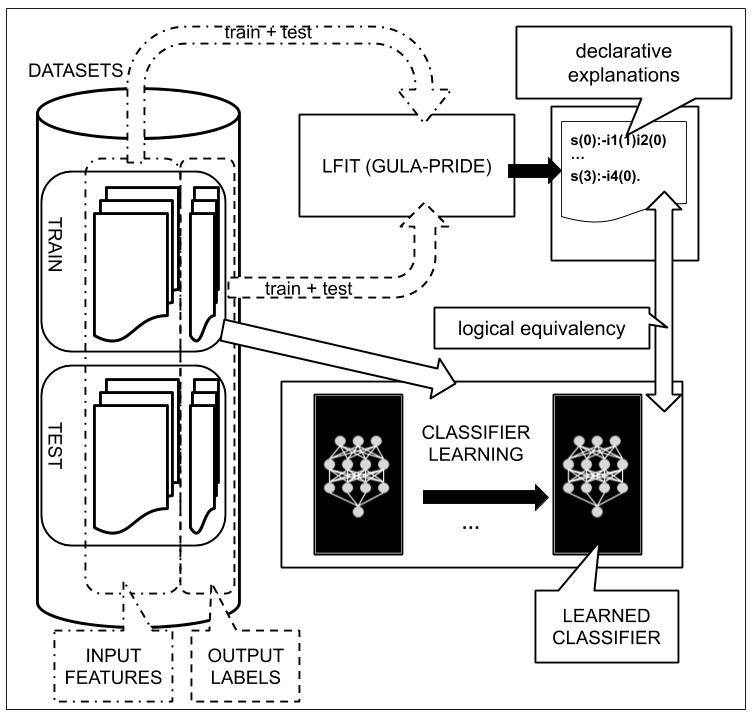}
\end{center}
   \caption{Experimental framework: \textbf{PRIDE} is fed with all the data available (train $+$ test) for increasing the accuracy of the equivalence. In our experiments we consider the classifier (see \cite{PenaAl20} for details) as a black box to perform regression from input resume attributes (atts.) to output labels (recruitment scores labelled by human resources experts). LFIT gets a digital twin to the neural network providing explainability (as human-readable white-box rules) to the neural network classifier.}
\label{fig:system_diagram}
\label{fig:onecol}
\end{figure}

Figure~\ref{fig:system_diagram} graphically describes our proposed approach to generate explanations using \textbf{LFIT} of a given black-box classifier. We can see there our purpose to get declarative explanations in parallel (in a kind of white-blox digital twin) to a given neural network classifier. In the present work, for our experiments we have used the same neural network and datasets described in \cite{PenaAl20} but excluding the face images as it is explained in the following sections.

\subsection{PRIDE Implementation of LFIT}

	{\bf GULA} \cite{TRILP18,TRMLJ2020} and {\bf PRIDE} \cite{ISTE2020} are particular implementations of the \textbf{LFIT} framework \cite{LFIT2013}. In the present section we introduce notation and describe the fundamentals of both methods.

	In the following, we denote by $\N := \{ 0, 1, 2, ... \}$
	the set of natural numbers,
	and for all $k, n \in \N$, $\segm{k}{n} := \{ i \in \N \mid k \leq i \leq n \}$
	is the set of natural numbers between $k$ and $n$ included.
	For any set $S$, the cardinal of $S$ is denoted $\card{S}$
	and the power set of $S$ is denoted $\powerset(S)$.
	
	Let $\V=\{\var_1,\dots,\var_n\}$ be a finite set of $n \in \N$ variables,
	$\Val$ the set in which variables take their values and
	$\dom : \V \rightarrow \powerset(\Val)$
	a function associating a domain to each variable.
	The atoms of \mvl\ (multi-valued logic) are of the form \vv\ where $\var\in\V$ and $\val\in\dom(\var)$.
    The set of such atoms is denoted by $\A^{\V}_{\dom} = \{\vv \in \V \times \Val \mid \val \in \dom(\var) \}$
	for a given set of variables $\V$
	and a given domain function $\dom$.
	In the following, we work on specific $\V$ and $\dom$
	that we omit to mention when the context makes no ambiguity,
	thus simply writing $\A$ for $\A^{\V}_{\dom}$.

	\begin{example}
		For a system of 3 variables, the typical set of variables is $\V = \{ a, b, c \}$.
		In general, $\Val = \N$ so that domains are sets of natural integers, for instance:
		$\dom(a) = \{ 0, 1 \}$,
		$\dom(b) = \{ 0, 1, 2 \}$ and
		$\dom(c) = \{ 0, 1, 2, 3 \}$.
		Thus, the set of all atoms is:
		$\A = \{ a^0, a^1, b^0, b^1, b^2, c^0, c^1, c^2, c^3 \}$.
	\end{example}

	A \mvl\ rule $R$ is defined by:
	\begin{equation}\label{discrete_rule}
	  R ~~=~~ \vvi{0} \leftarrow \vvi{1} \wedge \cdots \wedge \vvi{m}
	\end{equation}
	where $\forall i \in \segm{0}{m}, \vvi{i} \in \A$ are atoms in \mvl{} 
	so that every variable is mentioned at most once in the right-hand part:
	$\forall j,k \in \segm{1}{m}, j \neq k \Rightarrow \var_j \neq \var_k$.
	Intuitively, the rule $R$ has the following meaning: the variable $\var_0$ can take the value $\val_0$ in the next dynamical step if for each $i \in \segm {1}{m}$, variable $\var_i$ has value $\val_i$ in the current dynamical step.

	The atom on the left-hand side of the arrow is called the {\em head} of $R$ and is denoted $h(R) := \vvi{0}$.
	The notation $\hvar{R} := \var_0$ denotes the variable that occurs in $h(R)$.
	The conjunction on the right-hand side of the arrow is called the {\em body} of $R$, written $b(R)$
	and can be assimilated to the set $\{\vvi{1},\dots,\vvi{m}\}$;
	we thus use set operations such as $\in$ and $\cap$ on it.
	The notation $\bvar{R} := \{ \var_1, \cdots, \var_m \}$ denotes the set of variables that occurs in $b(R)$.
	More generally, for all set of atoms $X \subseteq \A$, we denote $\setvar{X} := \{ \var \in \V \mid \exists \val \in \dom(\var), \vv \in X \}$ the set of variables appearing in the atoms of $X$.
	A {\it multi-valued logic program} (\mvlp) is a set of \mvl\ rules.
	
	\pref{def:domination} introduces a domination relation between rules
	that defines a partial anti-symmetric ordering.
	Rules with the most general bodies dominate the other rules.
	In practice, these are the rules we are interested in since they cover the most general cases.

	\begin{definition}[Rule Domination]
	\label{def:domination}
		Let $R_1$, $R_2$ be two \mvl\ rules.
	  	The rule $R_1$ {\em dominates} $R_2$, written $R_2 \leq R_1$ if $h(R_1) = h(R_2)$ and $b(R_1)\subseteq b(R_2)$.
	\end{definition}
    
    In \cite{TRMLJ2020}, the set of variables is divided into two disjoint subsets: $\T$ (for targets) and $\F$ (for features).
    It allows to define dynamic \mvlp\ which capture the dynamics of the problem we tackle in this paper.
	
	\begin{definition}[Dynamic \mvlp]
		Let $\T \subset \V$ and $\F \subset \V$ such that $\F = \V \setminus \T$.
		A \dmvlp\ $P$ is a \mvlp\ such that $\forall R \in P,
		\hvar{R} \in \T$ and $\forall \vv \in b(R), \var \in \F$.
	\end{definition}

	The dynamical system we want to learn the rules of is represented by a succession of {\em states} as formally given by \pref{def:discrete_state}.
	We also define the “compatibility” of a rule with a state in \pref{def:matching}.
	
	\newcommand{\vvx}[1]{\ensuremath{\var_{#1}^{x_{#1}}}}

	\begin{definition}[Discrete state]
	\label{def:discrete_state}
	  A {\em discrete state} $s$ on $\T$ (\resp $\F$) of a \dmvlp\ is a function from $\T$ (\resp $\F$) to $\mathbb{N}$, \ie it associates an integer value to each variable in $\T$ (\resp $\F$).
	  It can be equivalently represented by the set of atoms
	  $\{ \var^{s(\var)} \mid \var \in \T\text{ (\resp \F)} \}$
	  and thus we can use classical set operations on it.
	  We write \SallT\ (\resp \SallF) to denote the set of all discrete states of $\T$ (\resp $\F$),
	  and a couple of states $(s,s') \in \SallF \times \SallT$ is called a \emph{transition}.
	\end{definition}
	
	\begin{definition}[Rule-state matching]
	\label{def:matching}
		Let $s \in \SallF$.
		The \mvl\ rule $R$ {\em matches} $s$, written $R\sqcap s$, if $b(R) \subseteq s$.
	\end{definition}
	
	The notion of transition in {\bf LFIT} correspond to a data sample in the problem we tackle in this paper: a couple features/targets.
	When a rule match a state it can be considered as a possible explanation to the corresponding observation.
	The final program we want to learn should both:
	\begin{itemize}
	    \item match the observations in a complete (all observations are explained) and correct (no spurious explanation) way;
	    \item represent only minimal necessary interactions (according to Occam's razor: no overly-complex bodies of rules)
	\end{itemize}
	{\bf GULA} \cite{TRILP18,TRMLJ2020} and {\bf PRIDE} \cite{ISTE2020} can produce such programs.

	Formally, given a set of observations $T$, {\bf GULA} \cite{TRILP18,TRMLJ2020} and {\bf PRIDE} \cite{ISTE2020} will learn a set of rules $P$ such that all observations are explained: $\forall (s,s') \in T, \forall \vv \in s', \exists R \in P, R \sqcap s, h(R) = \vv$.
	All rules of $P$ are correct w.r.t. $T$: $\forall R \in P, \forall (s1,s2) \in T, R \sqcap s1 \implies \exists (s1,s3) \in T, h(R) \in s3$ (if $T$ is deterministic, $s2 = s3$).
	All rules are minimal w.r.t. $\F$: $\forall R \in P, \forall R' \in \mvlp, R'$ correct w.r.t. $T$ it holds that $R \leq R' \implies R' = R$.
	
	The possible explanations of an observation are the rules that match the feature state of this observation.
	The body of the rules gives minimal condition over feature variables to obtain its conclusions over a target variable.
	Multiple rules can match the same feature state, thus multiple explanations can be possible.
	Rules can be weighted by the number of observations they match to assert their level of confidence.
    Output programs of {\bf GULA} and {\bf PRIDE} can also be used in order to predict and explain from unseen feature states by learning additional rules that encode when a target variable value is not possible as shown in the experiments of \cite{TRMLJ2020}.
	
	\newcommand{\U}{\ensuremath{\mathcal{U}}}
	\newcommand{\Up}{\ensuremath{\U_P}}

	\newcommand{\rrea}{\ensuremath{s\xrightarrow{R}s'}}
	\newcommand{\prea}{\ensuremath{s\xrightarrow{P}s'}}
	\newcommand{\preasp}{\ensuremath{s\xrightarrow{P}s''}}
	\newcommand{\rea}[1]{\ensuremath{s\xrightarrow{#1}s'}}
	\newcommand{\srea}[2]{\ensuremath{\xhookrightarrow{#2}{#1}}}

\section{Experimental Framework\label{experiments}}
\subsection{Dataset}

For testing the capability of \textbf{PRIDE} to generate explanations in machine learning domains we have designed several experiments using the FairCVdb dataset \cite{PenaAl20}. 

FairCVdb comprises $24$,$000$ synthetic resume profiles. Each resume includes $12$ features ($\var_i$) related to the candidate merits, $2$ demographic attributes (gender and three ethnicity groups), and a face photograph. In our experiments, we discarded the face image for simplicity (unstructured image data will be explored in future work). Each of the profiles includes three target scores ($T$) generated as a linear combination of the $12$ features: 

\begin{equation}
\label{eqn:Score_gen}
    T = \beta + \sum_{i = 1}^{12} \alpha_{i} \cdot \var_i, 
\end{equation}

\noindent where $\alpha_{i}$ is a weighting factor for each of the merits (see \cite{PenaAl20} for details): $i)$ unbiased score ($\beta=0$); $ii)$ gender-biased scores ($\beta=0.2$ for male and $\beta=0$ for female candidates); and $iii)$ ethnicity-biased scores ($\beta=0.0, 0.15$ and $0.3$ for candidates from ethnic groups 1, 2 and 3 respectively). Thus we intentionally introduce bias in the candidate scores.
From this point on we will simplify the name of the attributes considering $g$ for gender, $e$ for ethnic group and $i1$ to $i12$ for the rest of input attributes.
In addition to the bias previously introduced, some other random bias was introduced relating some attributes and gender to simulate real social dynamics. The attributes concerned were $i3$ and $i7$.
Note that merits were generated without bias, assuming an ideal scenario where candidate competencies do not depend on their gender of ethnic group. 
For the current work we have used only discrete values for each attribute discretizing one attribute (experience to take values from 0 to 5, the higher the better) and the scores (from 0 to 3) that were real valued in  \cite{PenaAl20}.


\subsection{Experimental Protocol: Towards Declarative Explanations}

We have experimented with \textbf{PRIDE} on the FairCVdb dataset described in the previous section.



\begin{figure}[t]
\begin{center}
        \includegraphics[width=1.0\linewidth]{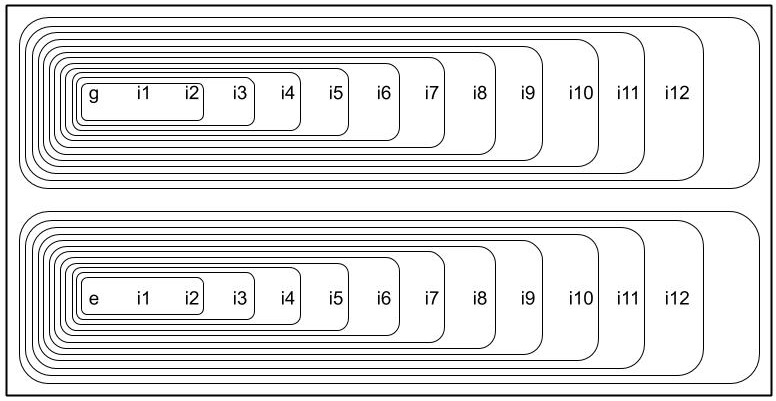}
\end{center}
   \caption{Structure of the experimental tests. There are 4 datasets for analysing gender (named $g$) and ethnicity ($e$) bias separately. Apart from gender and ethnicity there are 12 other input attributes (named from $i1$ to $i12$). There is a couple of (biased and unbiased) datasets for each one: gender and ethnicity. We have studied the input attributes by increasing complexity starting with $i1$ and $i2$ and adding one at each time. So, for each couple we have considered 11 different scenarios (named from $s1$ to $s11$). This figure shows their structure ($s_i$ is included in all $s_j$ for which $i < j$).}
\label{fig:scenarios}
\label{fig:onecol}
\end{figure}


Figure~\ref{fig:scenarios} shows names and explains the scenarios considered in our experiments. In \cite{PenaAl20}, researchers demonstrate that an automatic recruitment algorithm based on multimodal machine learning reproduces existing biases in the target functions even if demographic information was not available as input (see \cite{PenaAl20} for details). Our purpose in the experiments is to obtain a declarative explanation capable of revealing those biases.

\section{Results\label{results}}

\subsection{Example of Declarative Explanation}

Listing \ref{listing:explain_s(3)} shows a fragment generated with the proposed methods for scenario $s1$ for gender-biased scores. We have chosen a fragment that fully \textit{explains} how a CV is scored with the value 3 for scenario 1. Scenario 1 takes into account the input attributes gender, education and experience. The first clause (rule), for example, says that if the value of a CV for the attribute gender is 1 (female), for education is 5 (the highest), and for experience is 3, then this CV receives the highest score (3). 

The resulting explanation is a propositional logic fragment equivalent to the classifier for the data seen. It can be also understood as a set of rules with the same behavior. From the viewpoint of explainable AI, this resulting fragment can be understood by an expert in the domain and used to generate new knowledge about the scoring of CVs.

\begin{lstlisting}[language=Prolog, caption=Fragment of explanation for scoring 3, label=listing:explain_s(3)]

scores(3) :- gender(1), 
    education(5), 
    experience(3).
scores(3) :- education(4), 
    experience(3).
\end{lstlisting}

\subsection{Quantitative Results: Identifying Biases}

Our quantitative results are divided in two parts. The first part is based on the fact that, in the biased experiments, if $gender(0)$ appears more frequently than $gender(1)$ in the rules, then that would lead to higher scores for $gender(0)$. In the second quantitative experimental part we will show the influence of bias in the distribution of attributes.

\subsubsection{Biased attributes in rules}

We first define Partial Weight $PW$ as follows. For any program $P$ and two atoms $v_0^{val_0^i}$ and $v_1^{val_1^j}$, where $val_0^i\in val_0$ and $val_1^i \in val_1$, define $S = \forall R \in P \wedge v_0^{val_0^i} \in h(R)\wedge v_1^{val_1^j} \in b(R)$. Then we have: $PW_{v_{1}^{val_{1}^j}}(v_0^{val_0^0}) = |S|$. A more accurate $PW$ could be defined, for example, setting different weights for rules with different length. But for our purpose, the frequency is enough. In our analysis, the number of examples for compared scenarios are consistent.

Depending on $PW$, we define Global Weight $GW$ as follows. For any program $P$ and $v_{1}^{val_{1}^j}$, we have: $GW_{v_{1}^{val_{1}^j}} = \sum_{val_0^i\in val_0} PW_{v_1^{val_1^j }} (v_0^{val_0^i })\cdot val_0^i$. The $GW_{v_{1}^{val_{1}^j}}$ is a weighted addition of all the values of the output, and the weight, in our case, is the value of scores. 

This analysis was performed only on scenario $s11$, comparing unbiased and gender- and ethnicity-biased scores. We have observed a similar behavior of both parameters: Partial and Global Weights. In unbiased scenarios the distributions of the occurrences of each value could be considered statistically the same (between \textit{gender(0)} and \textit{gender(1)} and among \textit{ethnicity(0)}, \textit{ethnicity(1)} and \textit{ethnicity(2)}). Nevertheless in biased datasets the occurrences of \textit{gender(0)} and \textit{ethnic(0)} for higher scores is significantly higher. The maximum difference even triplicates the occurrences of the other values.

For the Global Weights, for example, the maximum differences in the number of occurrences, without and with bias respectively, for higher scores expressed as $\%$ increases from $48.8\%$ to $78.1\%$ for \textit{gender(0)} while for \textit{gender(1)} decreases from $51.2\%$ to $21.9\%$. In the case of \textit{ethnicity}, it increases from $33.4\%$ to $65.9\%$ for \textit{ethnic(0)}, but decreases from $33.7\%$ to $19.4\%$ for \textit{ethnic(1)} and from $32.9\%$ to $14.7\%$ for \textit{ethnic(2)}.

\subsubsection{Distribution of biased attributes}

We now define $freq_{p_1}(a)$ as the frequency of attribute $a$ in $P_1$. The normalized percentage for input $a$ is: $NP_{p_1}(a) = freq_{p_1}(a)/\sum_{x \in input} freq_{p_1}(x)$ and the percentage of the absolute increment for each input from unbiased experiments to its corresponding biased ones is defined as: $AIP_{p_1,p_2}(a) = ( freq_{p_1}(a)- freq_{p_2}(a))/ freq_{p_2}(a)$.

In this approach we have taken into account all the scenarios (from $s1$ to $s11$) for both \textit{gender} and \textit{ethnicity}.

\begin{figure}[t]
\begin{center}
        \includegraphics[width=\linewidth]{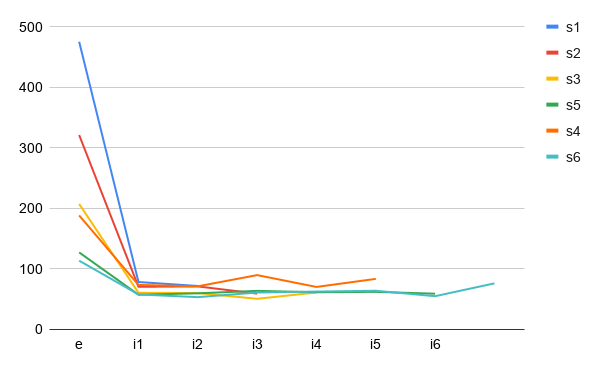}
\end{center}
   \caption{Percentage of the absolute increment (comparing scores with and without bias for ethnicity) of each attribute for scenarios s1, s2, s3, s4, s5 and s6 ($AIP_{us1-6,ebs1-6}$). The graphs link the points corresponding to all the input attributes considered in each scenario.}
\label{fig:e_graph_analysis_1}
\label{fig:onecol}
\end{figure}

\begin{figure}[t]
\begin{center}
        \includegraphics[width=\linewidth]{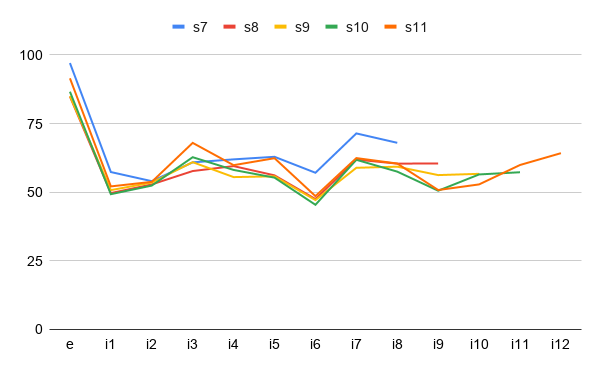}
\end{center}
   \caption{$AIP_{us7-11,ebs7-11}$}
\label{fig:e_graph_analysis_2}
\label{fig:onecol}
\end{figure}

We have observed that for both parameters the only attributes that consistently increase their values are \textit{gender} and \textit{ethnicity} comparing unbiased and gender/ethnicity-biased scores.
Figures~\ref{fig:e_graph_analysis_1} and \ref{fig:e_graph_analysis_2} show $AIP_{us1-11,ebs1-11}$ for each attribute, that is, their values comparing unbiased and ethnic-biased scores for all the scenarios from $s1$ to $s11$. It is clear that the highest values correspond to the attribute \textit{ethnicity}.

\begin{figure}[t]
\begin{center}
        \includegraphics[width=\linewidth]{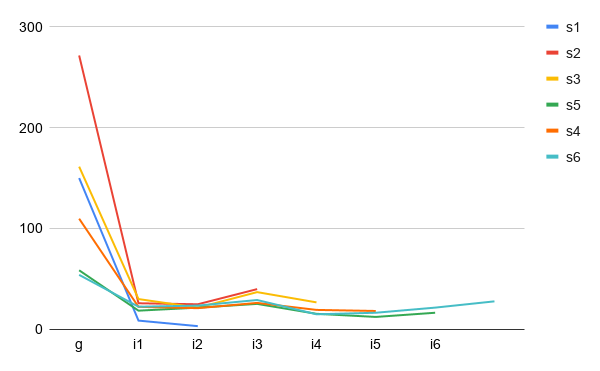}
\end{center}
   \caption{$AIP_{us1-6,ebs1-6}$}
\label{fig:g_graph_analysis_1}
\label{fig:onecol}
\end{figure}

\begin{figure}[t]
\begin{center}
        \includegraphics[width=\linewidth]{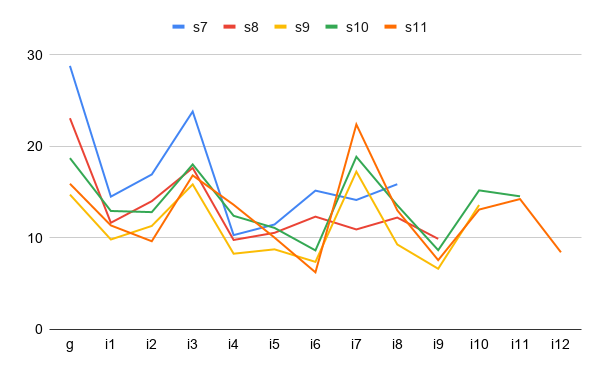}
\end{center}
   \caption{$AIP_{us7-11,gbs7-11}$}
\label{fig:g_graph_analysis_2}
\label{fig:onecol}
\end{figure}

\begin{figure}[t]
\begin{center}
        \includegraphics[width=\linewidth]{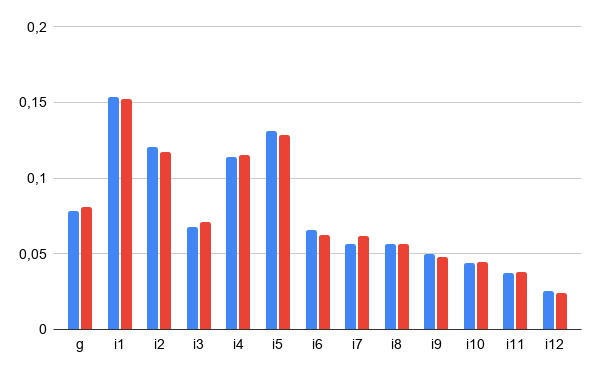}
\end{center}
   \caption{Normalized percentage of frequency in scenario s11 of each attribute: g, i1 to i11 ($NP_{s11}$). No bias (blue), Gender-biased scores (red).}
\label{fig:g_graph_analysis_3}
\label{fig:onecol}
\end{figure}

Something similar happens for gender. Figures~\ref{fig:g_graph_analysis_1} and \ref{fig:g_graph_analysis_2} show $AIP_{us1-11,gbs1-11}$ for each attribute when studying gender-biased scores. It is worth mentioning some differences in scenarios $s9$, $s10$ and $s11$ regarding attributes $i3$ and $i7$. These apparent anomalies are explained by the random bias introduced in the datasets in order to relate these attributes with gender when the score is biased.
Figure~\ref{fig:g_graph_analysis_3} shows $NP_{s11}$ for all the attributes. It clearly shows the small relevance of attributes $i3$ and $i7$ in the final biased score.
As it is highlighted elsewhere, this capability of \textbf{PRIDE} to identify this random indirect perturbation of other attributes in the bias is a relevant achievement of our proposal.

\section{Discussion\label{discussion}}


After running the experiments described in the previous sections we can extract the following conclusions.

\begin{itemize}

\item \textbf{PRIDE can \textit{explain} algorithms learnt by neural networks.} The theorems that support the characteristics of PRIDE allow \textit{to get a set of propositional clauses logically equivalent to the systems} observed when facing the input data provided. In addition, \textit{each proposition has a set of conditions that is minimum}. So, once the scorer is learnt, PRIDE translates it into a logical equivalent program. This program is a list of clauses like the one shown in Listing \ref{listing:explain_s(3)}. Logical programs are declarative theories that explain the knowledge on a domain. 


\item \textbf{PRIDE can \textit{explain} what happens in a specific domain.} Our experimental results discover these characteristics of the domain:

\begin{itemize}

\item \textit{\textbf{Insights into the structure of the datasets}}. We have seen (and further confirmed with the authors of the datasets) some characteristics of the datasets, e.g.: 1) \textit{All the attributes are needed for the score}. We have learnt the logical version of the system starting from only two input attributes and including one additional attribute at a time and we only reached an accuracy of 100 \% when taking into account all of them. This is because removing some attributes generates indistinguishable CVs (all the remainder attributes have the same value) with different scores (that correspond to different values in some of the removed attributes). 2) \textit{Gender and ethnicity are not the most relevant attributes for scoring}: The number of occurrences of these attributes is much smaller than others in the conditions of the clauses of the learnt logical program. 3) While trying to catch the biases we have discovered that \textit{some attributes seem to increase their relevance when the score is biased}. For example, the competence in some specific languages (attribute i7) seems to be more relevant when the score has gender bias. 
After discussing with the authors of the datasets, they confirmed a random perturbation of these languages into the biases, that explained our observations. 

\item \textbf{\textit{Biases in the training datasets are detected.}} We have analysed the relationship between the scores and the specific values of the attributes used to generated the biased data. We have proposed a simple mathematical model based on the \textit{effective weights} of the attributes that concludes that higher values of the scores correspond to the same specific values of gender (for gender bias) and ethnic group (for ethnicity bias).
On the other hand, we have performed an exhaustive series of experiments to analyse the increase of the presence of the gender and ethnicity in the conditions of the clauses of the learnt logical program (comparing the unbiased and biased versions). 

\end{itemize}

\end{itemize}

Our overall conclusion is that 
\textbf{LFIT}, and in particular \textbf{PRIDE}, is able to offer explanations to the algorithm learnt in the domain under consideration. The resulting explanation is, as well, expressive enough to catch training biases in the models learnt with neural networks.

\section{Conclusions\label{conclusions}}

The main goal of this paper was to check if ILP (and more specifically LFIT with \textbf{PRIDE}) could be useful to provide \textit{declarative explanations} in machine learning by neural networks.

The domain selected for our experiments in this first entry to the topic is one in which the explanations of the learned models' outputs are specially relevant: automatic recruitment algorithms. In this domain, ethic behavior is needed, no spurious biases are allowed. For this purpose, a pack of synthetically generated datasets has been used. The datasets contain resumes (CVs) used in \cite{PenaAl20} for testing the ability of deep learning approaches to reproduce and remove biases present in the training datasets. In the present work, different input attributes (including the resume owner merits, gender, and ethnicity) are used to score each CV automatically using a neural network. Different setups are considered to introduce artificial gender- and ethnicity-bias in the learning process of the neural network. In \cite{PenaAl20} face images were also used and the relationship between these pictures and the biases was studied (it seems clear that from the face you should be able to deduce the gender and ethnic group of a person). Here we have removed images because PRIDE is more efficient with pure discrete information.

Our main goal indicated above translates into these two questions: Is PRIDE expressive enough to explain how the program learnt by deep-learning approaches works? Does PRIDE catch biases in the deep-learning processes?
We have given positive answer to both questions.

\section{Further Research Lines\label{further}}


\begin{itemize}

\item \textbf{Increasing understandability.}
Two possibilities could be considered in the future:
$1)$ to \textit{ad hoc} post-process the learnt program for translating it into a more abstract form, or $2)$ to increase the expressive power of the formal model that supports the learning engine using, for example, ILP based on first order logic.

\item \textbf{Adding predictive capability.} 
\textbf{PRIDE} 
is actually not aimed to predict but to explain (declaratively) by means of a digital twin of the observed systems. Nevertheless, it is not really complicated to extend \textbf{PRIDE} functionality to predict. It should be necessary to change the way in which the result is interpreted as a logical program: 
mainly by adding mechanisms to chose the most promising rule when more than one is applicable.

Our plan is to test an extended-to-predict \textbf{PRIDE} version to this same domain and compare the result with the classifier generated by deep learning algorithms.

\item \textbf{Handling numerical inputs.} 
\cite{PenaAl20} included as input the images of the faces of the owners of the CVs. Although some variants to \textbf{PRIDE} are able to cope with numerical signals, the huge amount of information associated with images implies performance problems. Images are a typical input format in real deep learning domains. We would like to add some automatic pre-processing step for extracting discrete information (such as semantic labels) from input images. 
We are motivated by the success of systems with similar approaches but different structure like \cite{AlirezaAl20}.

\item \textbf{Measuring the accuracy and performance of the explanations.} 
As far as the authors know there is no standard procedure to evaluate and compare different explainability approaches. We will incorporate in future versions some formal metric.
\end{itemize}

\section{Acknowledgements}

This work has been supported by projects: PRIMA (H2020-MSCA-ITN-2019-860315), TRESPASS-ETN (H2020-MSCA-ITN-2019-860813), IDEA-FAST (IMI2-2018-15-853981), BIBECA (RTI2018-101248-B-I00 MINECO/FEDER), RTI2018-095232-B-C22 MINECO, and Accenture.


{\small
\bibliographystyle{ieee_fullname}
\bibliography{pride_XAI}
}

\end{document}